\documentclass[10pt,twocolumn,letterpaper]{article}

\usepackage[pagenumbers]{cvpr} % To force page numbers, e.g. for an arXiv version

\usepackage{arydshln}
\definecolor{cvprblue}{rgb}{0.21,0.49,0.74}
\definecolor{roma}{rgb}{0.7, 0, 0}
\definecolor{aerialmast3r}{rgb}{0, 0, 0.7}
\definecolor{spider}{rgb}{0, 0.7, 0}

\definecolor{best}{rgb}{0.96, 0.57, 0.58}
\definecolor{second}{rgb}{0.98, 0.78, 0.57}
\usepackage{colortbl}
\usepackage[pagebackref,breaklinks,colorlinks,allcolors=cvprblue]{hyperref}
\usepackage{booktabs} 
\usepackage{xcolor} 
\usepackage{multirow} 
\usepackage{graphicx}
\usepackage{gensymb}
\usepackage{adjustbox}

\title{SPIDER: Spatial Image CorresponDence Estimator for Robust Calibration}

\author{
Zhimin Shao \quad
Abhay Yadav \quad
Rama Chellappa \quad
Cheng Peng \\[0.3em]
Johns Hopkins University, Baltimore, USA \\[0.2em]
{\tt\small \{zshao14, ayadav13, rchella4, cpeng26\}@jhu.edu} \\
\href{https://zhimin00.github.io/spider.github.io/}{
\adjustbox{}{\includegraphics[height=1em]{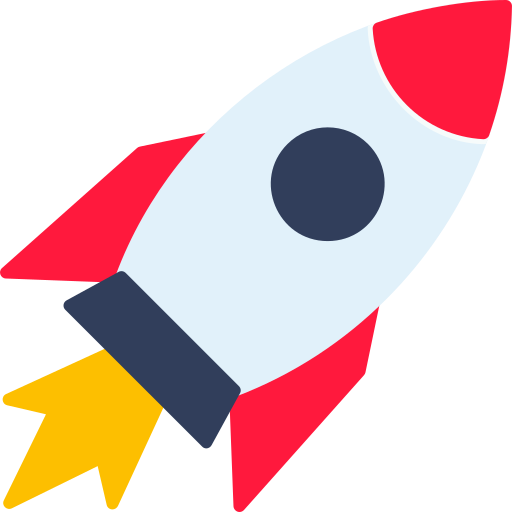}}~
Project Page}
\quad
\href{https://github.com/Zhimin00/spider/}{
\adjustbox{}{\includegraphics[height=1em]{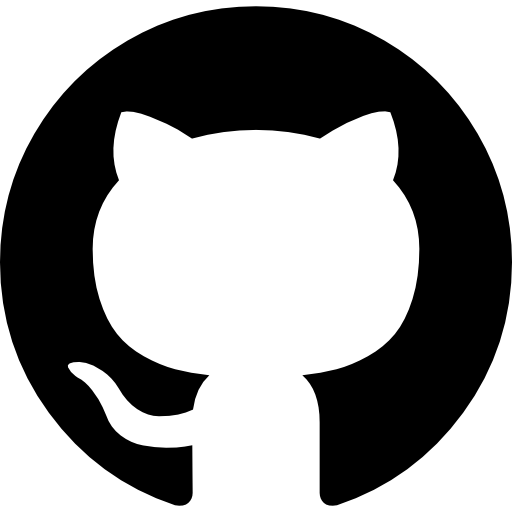}}~GitHub}
}

\begin{document}

\twocolumn[{
\renewcommand\twocolumn[1][]{#1}
\maketitle
\vspace{-2em}
\begin{center}
    \includegraphics[width=\linewidth]{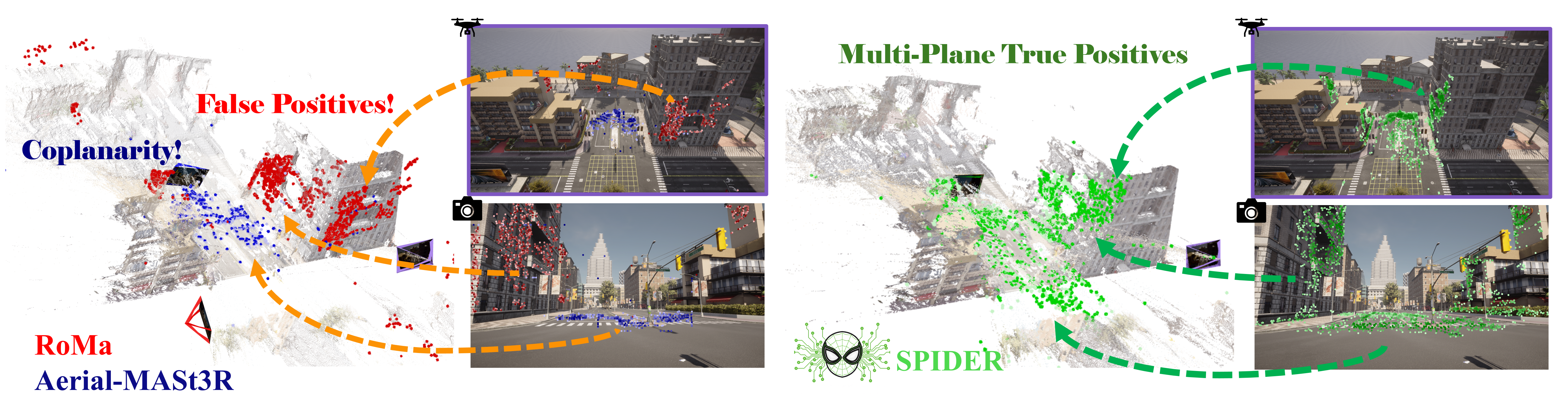}
    \vspace{-1.5em}
    \captionof{figure}
    {% \textbf{Dense Correspondences.}
    Visualized in 3D, 
\textcolor{spider}{SPIDER} jointly predicts pixel-wise warps and feature descriptors even across large viewpoint changes, unifying the appearance sensitivity of \textcolor{roma}{RoMa} and the geometric consistency of \textcolor{aerialmast3r}{Aerial-MASt3R} within a single framework. This enables accurate camera calibration and pose estimation, achieving and surpassing State-Of-the-Art performance on challenging benchmarks.}
    \label{fig:teaser}
\end{center}
}]

\begin{abstract}
Reliable image correspondences form the foundation of vision-based spatial perception, enabling recovery of 3D structure and camera poses. However, unconstrained feature matching across domains such as aerial, indoor, and outdoor scenes remains challenging due to large variations in appearance, scale and viewpoint. Feature matching has been conventionally formulated as a 2D-to-2D problem; however, recent 3D foundation models provides spatial feature matching properties based on two-view geometry. While powerful, we observe that these spatially coherent matches often concentrate on dominant planar regions, e.g., walls or ground surfaces, while being less sensitive to fine-grained geometric details, particularly under large viewpoint changes.
To better understand these trade-offs, we first perform linear probe experiments to evaluate the performance of various vision foundation models for image matching. Building on these insights, we introduce \textbf{SPIDER}, a universal feature matching framework that integrates a shared feature extraction backbone with two specialized network heads for estimating both 2D-based and 3D-based correspondences from coarse to fine. Finally, we introduce an image-matching evaluation benchmark that focuses on unconstrained scenarios with large baselines. SPIDER significantly outperforms SoTA methods, demonstrating its strong ability as a universal image-matching method. 
\end{abstract}    
\vspace{-3em}
\section{Introduction}
 Image matching is a fundamental problem in Computer Vision, encompassing camera calibration~\cite{schoenberger2016sfm}, novel view synthesis~\cite{mildenhall2021nerf, kerbl20233d}, 3D reconstruction~\cite{schoenberger2016mvs, wang2021neus}, image localization~\cite{sarlin2019coarse}, etc. The goal of image matching is to estimate correspondences between a image pair that refer to the same 3D points, such that 3D geometry can be estimated.

Recent dense matching approaches focus on estimating pixel-to-pixel correspondences, with methods ranging from supervised learning to leveraging large-scale pretrained vision models. Earlier approaches rely on learned dense features~\cite{sun2021loftr,chen2022aspanformer,sarlin2020superglue,truong2023pdc,edstedt2023dkm}, which improve matching accuracy but struggle to generalize to unseen scenarios. With the rise of large-scale visual data and self-supervised learning, frozen 2D-pretrained backbones trained on representation learning or image reconstruction tasks~\cite{rombach2022high,oquab2023dinov2,simeoni2025dinov3} have exhibited geometric and semantic alignment capabilities~\cite{el2024probing,tang2023emergent}. Such representations have also been leveraged for image matching. For example, RoMa~\cite{edstedt2024roma} employs DINOv2~\cite{oquab2023dinov2} to estimate dense pixel-to-pixel warps. At their core, these methods rely on appearance similarity, whether hand-crafted, learned or 2D-pretrained features. As shown in \cref{fig:teaser}, one downside for such an approach is its propensity to match semantically similar but geometrically inconsistent regions, e.g., windows across different walls.

3D-pretrained vision models have emerged as a new class of Vision Foundation Models (VFMs) that directly decode 3D geometry. These methods can be broadly categorized into two types. Single-view 3D models estimate monocular depth from large-scale datasets, enabling implicit 3D awareness from a single image~\cite{bochkovskii2024depth,yang2024depth}. In contrast, multi-view 3D models are trained on image pairs or sets with geometric objectives such as cross-view completion, point cloud prediction, etc.~\cite{weinzaepfel2022croco,wang2024dust3r,wang2025vggt}. 
% These models learn geometry-consistent features. 
MASt3R~\cite{leroy2024grounding} and Aerial-MASt3R~\cite{vuong2025aerialmegadepth} extend upon this paradigm by explicitly predicting correspondences, resulting in matches that follow spatial patterns rather than 2D edges or corners. This property allows 3D VFMs to more reliably match sparse or textureless regions% and achieve high specificity in correspondence estimation
. However, as shown in \cref{fig:teaser}, their sensitivity remains limited due to challenges in estimating feasible depth under wide baselines. Moreover, the characteristics and transferability of these pretrained 3D features remain largely unexplored in image matching.

In real-world scenarios where image pairs can have large baselines, low overlap, or varying aquisition conditions, two key challenges emerge for constructing a universal feature matching model.
First, the model must achieve high sensitivity, i.e., the ability to reliably detect correspondences across appearance, scale, and viewpoint variations.
Second, it must maintain high specificity, i.e., the ability to reject visually similar but geometrically inconsistent regions. Despite recent advances, a dilemma still exists: pattern-driven methods are sensitive but not specific, and geometry-driven methods are specific but not sensitive.

To this end, we present \textbf{SPIDER}, a Spatial Image corresponDence Estimator for Robust calibration. We revisit dense correspondence estimation from a coarse-to-fine perspective. Our first key idea is to employ a 3D VFM as the unified feature extraction backbone, which provides consistent geometry-aware representations across views.
While 3D VFMs offer strong geometric priors, their representations often lack the fine-grained details required for wide-baseline matching. We attribute this limitation to the reliance of coarse geometry cues learned during Vision Transformer patchification, which can suppress local appearance information.
% crucial for pixel-level alignment. 
To recover these details, SPIDER introduces 2D ConvNet encoders as a fine-level branch to capture appearance-sensitive features. 

To further balance sensitivity and specificity, SPIDER employs dual correspondence heads, called Spatial-Image Matcher (SIM). Both heads 
% follow a multi-scale coarse-to-fine formulation, 
leverage the hierarchical features extracted by the coarse 3D VFM and the fine 2D ConvNet encoders. The pattern-driven warp head performs continuous pixel-level regression, providing high sensitivity to local 
% geometric 
appearance variations and sub-pixel alignment. In contrast, the geometry-driven feature descriptor produces discriminative embeddings for mutual nearest-neighbor matching, ensuring high specificity by rejecting visually ambiguous correspondences. 
% Rather than jointly optimizing the two heads, SPIDER learns them independently and fuses their correspondences at inference time. 
The two match heads are synergistic and their fusion lead to better calibration accuracy and overall matching robustness.

Extensive experiments show that SPIDER achieves State-of-The-Art performance on a variety of image matching benchmarks~\cite{shen2024gim}, demonstrating its balance between geometric consistency and appearance robustness.
We further evaluate SPIDER under unconstrained scenarios, including cross-elevation and wide-baseline settings, where it consistently surpasses prior methods. 
% —establishing it as a universal and robust feature matching framework for real-world applications.

To summarize, our main contributions are:
\begin{itemize}
\item We introduce SPIDER, an unconstrained feature matching algorithm based on a 3D VFM + ConvNet combination as a multi-scale feature extraction backbone. 
\item We introduce a dual-head Coarse-to-Fine matching head, which produces feature descriptors and warp estimations for geometry-based and pattern-based matches.
% \item We propose an unconstrained feature matching benchmark to analyze the tradeoff between sensitivity and specificity amongst current methods.
\item We conduct extensive experiments, and find SPIDER to excel in all challenging scenarios and achieve a balance between high sensitivity and specificity, particularly in unconstrained environments.
\end{itemize}

%-------------------------------------------------------------------------

\section{Related Works}
\textbf{Image Matching.} Image matching methods have evolved from sparse to detector-free and dense formulations~\cite{ma2021image,zhang2025deep}. Early approaches~\cite{lowe2004distinctive, bay2006surf, rublee2011orb} rely on sparse keypoint detection and description.
% , with classical hand-crafted methods
These methods can be further improved by learnable approaches~\cite{detone2018superpoint,truong2023pdc}. Detector-free approaches~\cite{chen2022aspanformer,sun2021loftr,tang2022quadtree, truong2023pdc} remove explicit keypoint detection and directly operate on coarse feature maps to establish correspondences, followed by mutual nearest-neighbor filtering and refinement. Recent 3D VFMs~\cite{leroy2024grounding,vuong2025aerialmegadepth} also incorporate a feature matching head and fall into this category. Building further, dense approaches including DKM~\cite{edstedt2023dkm} and RoMa~\cite{edstedt2024roma} compute pixel-wise or nearly pixel-wise correspondences, typically via dense warp estimation, enabling accurate matches without relying on explicit keypoints. More recently, VGGT~\cite{wang2025vggt} has integrated geometric priors into transformer-based representations, enabling joint prediction of camera poses, depth maps, pointmaps, and tracking in a single feed-forward pass.

\noindent\textbf{Vision Foundation Models.} 
Recent years have witnessed rapid progress in VFMs, which learn transferable representations through diverse data sources and training paradigms. 
Early 2D backbones such as VGG-19~\cite{simonyan2014very} and ResNet-50~\cite{he2016deep} are trained on ImageNet using full supervision. 
Self-supervised and Masked Image Modeling methods such as  DINOv2~\cite{oquab2023dinov2} and its successor DINOv3~\cite{simeoni2025dinov3} dramatically scale data and compute to produce highly semantic, domain-general features. Stable Diffusion extend this trend by combining contrastive, generative, and reconstruction objectives, while AM-RADIO~\cite{ranzinger2024radio} demonstrates that multiple VFMs can be unified through multi-teacher distillation. They have become widely applied for dense vision tasks by providing strong appearance priors while mitigating overfitting~\cite{kappeler2024few, li2025meddinov3}. 
However, despite their strong semantic alignment, these 2D models lack explicit understanding geometry, depth, or camera pose.
% remain fundamentally single-view.

% : they lack explicit understanding geometry, depth, or camera pose.

In contrast, 3D VFMs learn from multi-view supervision, directly encoding geometric consistency across images.
CroCo~\cite{weinzaepfel2022croco} introduces cross-view completion as a pretext task, leveraging paired views to reconstruct one image from another, thereby coupling appearance and structure. Building on this, DUSt3R~\cite{wang2024dust3r} formulates 3D reconstruction problem as a regression of pointmaps and implicitly learns view transformations through cross-attention between image pairs. MASt3R~\cite{leroy2024grounding} extends this idea by integrating a new head that provides features aligned with metric scene geometry for matching. Most recently, VGGT~\cite{wang2025vggt} generalizes 3D pretraining to a broader range of tasks, serving as a unified 3D VFM backbone.

% These 3D-supervised models encode spatial relationships and depth priors that substantially improve downstream tasks such as image matching, camera localization, and reconstruction, motivating our choice to build upon them in SPIDER.

\section{SPIDER}
As a fundamental task that underpins many downstream computer vision algorithms, feature matching remains difficult in unconstrained environments.
SPIDER seeks to establish accurate and dense correspondences across two views, even if they are from \textit{wide baselines} or \textit{visually ambiguous}. At a high level, two components provide SPIDER the ability to achieve this goal: 1. a strong, geometry-aware feature extraction backbone augmented by high resolution convolution features, and 2. an ensemble of pattern-driven and geometry-driven correspondence estimation heads.

% To jointly achieve semantic diversity and geometric consistency, we propose SPIDER, a dual-head coarse-to-fine matching architecture built upon 3D-pretrained foundation features. An overview of the framework is shown in \cref{fig:spider}. We first review the 3D foundation backbone in \cref{s3.1}, followed by our proposed matching heads in \cref{s3.2}, training objective in \cref{s3.3} and training strategy in \cref{s3.4}.

\begin{figure*}
    \centering
    \includegraphics[width=1\linewidth]{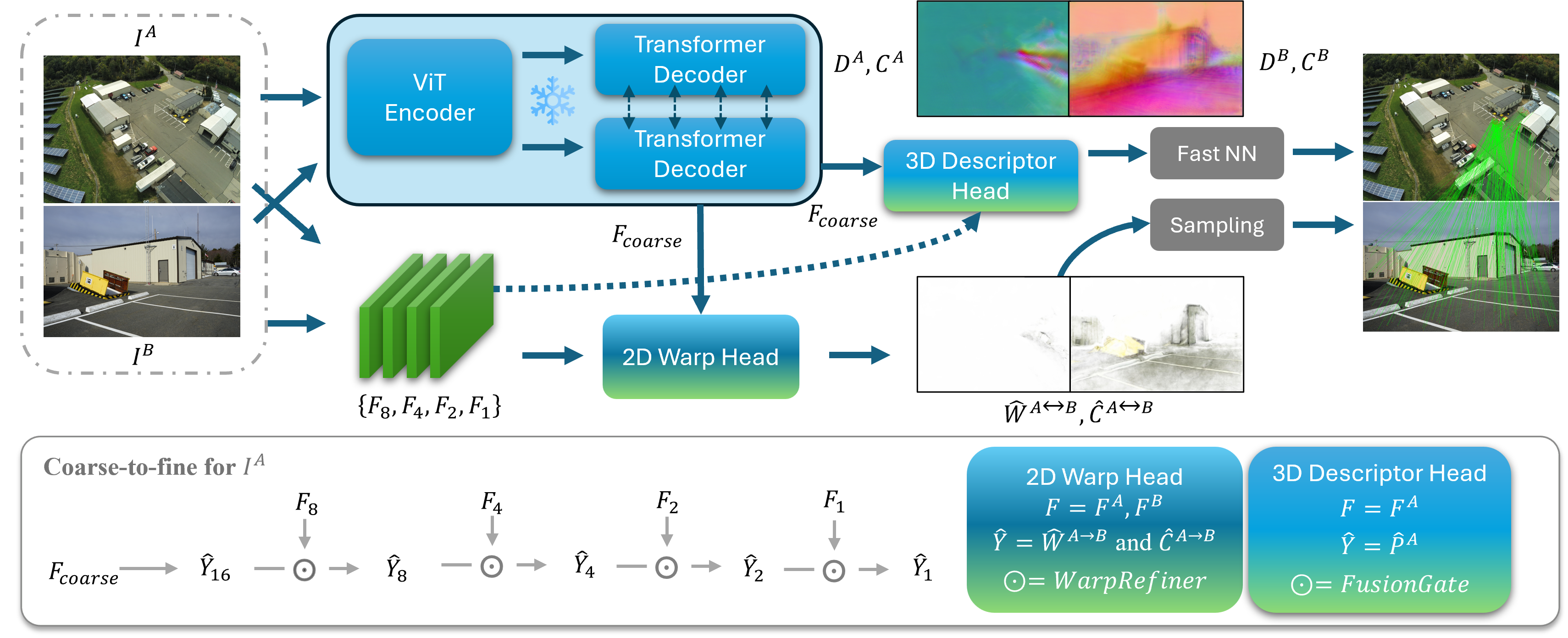}
    \caption{\textbf{Method Overview.} Given two input images $I^A$ and $I^B$, our method builds on 3D VFM features and ConvNet features to combine semantic alignment and geometric consistency. A dual-head architecture operates in a coarse-to-fine manner: (1) the descriptor head aggregates multi-scale features through attention-based Fusion Gates to produce geometry-aware descriptors and confidence maps; (2) the warp head predicts dense correspondence fields and confidence maps, progressively refined across multiple scales. Final correspondences are sampled from the predicted warp and fastNN. }
    \label{fig:spider}
\end{figure*}
\subsection{Multi-Scale 3D Vision Foundation Models}\label{s3.1}
Given two input images $I^A$ and $I^B$, SPIDER first perform feature extraction through a neural network $\mathcal{M}_c$:
\begin{equation}
  F^A_{\text{coarse}}, F^B_{\text{coarse}} = \mathcal{M}_c(I^A, I^B),
  \label{eq:fm}
\end{equation}
While previous methods~\cite{edstedt2024roma, leroy2024grounding} have shown that both 2D and 3D VFMs are helpful as the feature extractor backbone for correspondence estimation, the optimal VFM choice remains an open question. To this end, we perform a large-scale evaluation of existing 2D and 3D VFMs by computing the pixel-wise feature distances between $\{F^A_{\text{coarse}}, F^B_{\text{coarse}}\}$ on groundtruth correspondences, i.e., linear probing, \emph{without} finetuning on the feature matching task or using the feature matching head.

As shown in \cref{tab:2dvs3d}, while 2D VFMs can achieve good EPE~\cite{oquab2023dinov2, li2025meddinov3} as an image encoder, 3D VFMs significantly improve spatial feature similarity and accuracy even though they are exposed to much less data. This is particularly true for DUSt3R~\cite{wang2024dust3r}, whose backbone is not trained for explicit matching. This demonstrates that the cross-attention decoders employed by common 3D VFMs are very effective.

Most recent VFMs adopt the Vision Transformer (ViT)~\cite{dosovitskiy2020vit} architecture~\cite{oquab2023dinov2, li2025meddinov3, leroy2024grounding, vuong2025aerialmegadepth,wang2025vggt} and share the patchification process. Each input image is divided into non-overlapping $16{\times}16$ patches for patch embedding and transformer encoding. Current 3D VFMs~\cite{leroy2024grounding, vuong2025aerialmegadepth,wang2025vggt} recover this resolution through feature upsample~\cite{shi2016real}; however, such upsampling still leads to smooth interpolation, as shown in~\cref{dust3r_f1} and~\cref{dust3r_f2}. To address this, SPIDER performs a second, multi-scale feature extraction through $\mathcal{M}_f$:
\begin{equation}
\begin{split}
  F^A_{\text{fine}}, F^B_{\text{fine}} = \mathcal{M}_f(I^A, I^B),\\
  F^v_{\text{fine}} = \{F^v_8, F^v_4, F^v_2, F^v_1\}, v\in \{A,B\}.
\end{split}
\end{equation}
The design of $\mathcal{M}_f$ is based on an efficient VGG~\cite{simonyan2014very}, allowing high resolution features from $I^A, I^B$ to directly shuffle to the matching heads.

\begin{table}
\centering
\caption{Linear-probe comparison between 2D- and 3D-pretrained backbones on the image-matching task. Lower EPE and higher PCK@32 indicate better performance.}
\resizebox{\linewidth}{!}{
\begin{tabular}{llcccc}
\toprule
\multirow{2}{*}{\textbf{Type}} & \multirow{2}{*}{\textbf{Method}} &  \multicolumn{2}{c}{\textbf{MegaDepth}} & \multicolumn{2}{c}{\textbf{Aerial-MegaDepth}}  \\
\cmidrule{3-6}
 &   &  \textbf{EPE} $\downarrow$ & \textbf{PCK@32} $\uparrow$ & \textbf{EPE} $\downarrow$ & \textbf{PCK@32} $\uparrow$   \\
\midrule
\multirow{7}{*}{\textbf{2D}} 
  & VGG-19~\cite{simonyan2014very} & 87.1  & 33.0\% & 118.1 & 17.8\% \\
  & ResNet-50~\cite{he2016deep}   & 66.6  & 42.8\% & 94.2 & 27.3\% \\
  & StableDiffusion~\cite{rombach2022high}   & 30.8  & 72.9\% & 78.8 & 39.3\%  \\
  & DINOv2~\cite{oquab2023dinov2}            & 24.9  & 77.7\% & 60.1 & 49.1\% \\
  & AM-RADIO~\cite{ranzinger2024radio}          & 47.8  & 51.8\% & 103.0 & 15.5\%   \\

  & DINOv3 (ViT-L)~\cite{simeoni2025dinov3}    & 17.6  & 88.1\% & 54.9 & 56.9\% \\
  & DINOv3 (ViT-7B)~\cite{simeoni2025dinov3}   & 15.8  & 89.6\% & 50.3 & 60.4\% \\
  
\midrule
\multirow{8}{*}{\textbf{3D}} 
  & DUNE ~\cite{sariyildiz2025dune}               & 27.1 & 74.9\% & 64.0 & 48.7\% \\
  & DUSt3R (enc.)~\cite{wang2024dust3r}           & 24.8  & 79.5\% & 69.8 & 44.1\% \\
  & DUSt3R (dec.)~\cite{wang2024dust3r}                  & 11.5  & 93.9\% & 37.4 & 71.3\%\\
  & MASt3R (enc.)~\cite{leroy2024grounding}            & 23.0  & 80.8\% 
 & 66.7 & 46.2\%  \\
  & MASt3R (dec.)~\cite{leroy2024grounding}                  & 10.2  & 94.8\% & 41.9 & 67.8\% \\
  & Aerial-MASt3R (enc.)~\cite{vuong2025aerialmegadepth}     & 21.7  & 83.3\% &52.7 & 56.4\% \\
  & \textbf{Aerial-MASt3R (dec.)}~\cite{vuong2025aerialmegadepth}  & \cellcolor{best}8.3   & \cellcolor{best}96.6\% & \cellcolor{best}28.0 & \cellcolor{best}77.8\%  \\
  & VGGT (enc.)~\cite{wang2025vggt} & 19.0  & 84.1\% & 55.8 & 55.5\%  \\
  & VGGT (dec.)~\cite{wang2025vggt} & \cellcolor{second}9.2  & \cellcolor{second}96.1\% & \cellcolor{second}32.6 & \cellcolor{second}75.8\% \\
\bottomrule
\end{tabular}
}
\label{tab:2dvs3d}
\end{table}
% The encoder extracts per-view representations, while the decoder performs cross-attention between multiple views, allowing information from one image to inform the representation of the other. We concatenate the final-layer encoder and decoder features to form the coarse representation for our matching heads. 
% As shown in \cref{s4.1}, the choice of backbone has a strong influence on downstream matching accuracy.
% % —geometry-supervised 3D backbones consistently outperform purely appearance-driven 2D variants.
% {\color{red}Explanation is needed here on why the selected backbone produces geometry aware descriptor (loss function based on depth), and why that's fundamentally different from 2D matcher}

\subsection{Spatial Image Matcher}\label{s3.2}
What is the best formulation for image matching once expressive features are extracted? As shown in \cref{fig:teaser}, 2D pattern-based and 3D geometry-based matching each has its pros and cons in unconstrained scenarios. SPIDER proposes a Spatial Image Matcher (SIM), which combines a dense descriptor head that learns geometry-aware features and a dense warp head that directly predicts pixel-wise correspondences. We detail how these heads are integrated with multi-scale features $F^A_{\text{coarse}}, F^B_{\text{coarse}},F^A_{\text{fine}}, F^B_{\text{fine}}$ below.

 % SPIDER comprises of two complementary heads for matching: a dense warp head that directly predicts pixel-wise correspondences between pairs, and a dense feature head that learns geometry-aware descriptors for correspondence verification. Both heads operate hierarchically from coarse to fine spatial scales, integrating global context from high-level ViT features and injecting fine detail through convolutional refinement.

\paragraph{Multi-Scale Descriptor Head.}
% The feature head refines hierarchical features from coarse to fine to generate dense per-pixel descriptors and confidence maps. 
% We also extract fine-scale features 
% $\{F^{v}_{\text{fine}}\}^\text{feat} = \{F^v_8, F^v_4, F^v_2, F^v_1\}$ 
% from each input image using another ConvNet encoder:
% \begin{equation}
% \{F_{\text{fine}}^v\}^\text{feat} = \text{ConvNet}_{\text{feat}}(I^v), v\in \{A,B\}
% \end{equation}
SPIDER first uses a series of neural networks $\mathcal{Q}_s$ to project multi-scale features to the same 128-dimensional in channel space: 
\begin{equation}
\begin{split}
    \hat{F}_\text{coarse}^v =Q_{16}(F_\text{coarse}^v), \hat{F}_\text{s}^v =Q_{s}(F_\text{s}^v),\\ v\in \{A,B\}, s\in {1,2,4,8}.
\end{split}
\end{equation}

For each view, we first estimate coarse feature descriptors $\hat{P}_{16}$ based on $F_\text{coarse}$ and a two-layer MLP $\mathcal{G}_{16}$:
\begin{equation}
    \hat{P}_{16}^v =\mathcal{G}_{16}(\hat{F}_\text{coarse}^v), v\in \{A,B\}.
\end{equation}
$\hat{P}_{s}$ is progressively upsampled by fusing with finer-scale features $\hat{F}_{\frac{s}{2}}$ 
through Fusion Gate modules. Specifically, $\hat{P}_s$ is first linearly upsampled and concatenated with $\hat{F}_{\frac{s}{2}}$. The combined features go through $\mathcal{G}_{\frac{s}{2}}$ to produce a spatial attention map~$\alpha$. 
This attention map controls the blending between $\hat{P}_s$ and $\hat{F}_{\frac{s}{2}}$, 
followed by a $3{\times}3$ convolution for local refinement. This process can summarized as:
\begin{equation}
\alpha_{\frac{s}{2}} = \sigma(\mathcal{G}_{\frac{s}{2}}([\uparrow_2(\hat{P}_s), \hat{F}_{\frac{s}{2}}])),
\end{equation}
\begin{equation}
y_{\frac{s}{2}} = \alpha_{\frac{s}{2}} \cdot \uparrow_2(\hat{P}_s) + (1 - \alpha_{\frac{s}{2}}) \cdot \hat{F}_{\frac{s}{2}},
\end{equation}
\begin{equation}
\hat{P}_{\frac{s}{2}} = \text{Conv}_{3\times3}(y_{\frac{s}{2}}),
\end{equation}
where $\sigma(\cdot)$ denotes the sigmoid function and $\uparrow_2$ denotes 2X linear interpolation.
This operation is applied sequentially from coarse to fine scales, i.e.,
\begin{equation}
\begin{split}
\hat{P}_{\frac{s}{2}}^v &= 
\text{FusionGate}_{s\rightarrow \frac{s}{2}}(\hat{P}^v_s, \hat{F}^v_{\frac{s}{2}}),\\
&\quad s \in \{16, 8, 4, 2\},\quad v\in \{A,B\}.
\label{eq:descriptor}
\end{split}
\end{equation}

This gating approach allows targeted changes to $\hat{P}_{16}$, which is already fairly accurate based on~\cref{tab:2dvs3d}. After the final fusion, the refined feature map $\hat{P}^v_{1}$ is passed through two lightweight prediction heads to produce dense feature descriptors and their confidence maps:
\begin{equation}
\hat{D}^v = \text{head}_{\text{desc}}(\hat{P}_1^v), 
\hat{C}^v = \text{head}_{\text{conf}}(\hat{P}_1^v), v\in \{A,B\}.
\end{equation}

\begin{figure}[]
    \setlength{\abovecaptionskip}{3pt}
    \setlength{\tabcolsep}{2pt}
    \begin{tabular}[b]{cc}
        \begin{subfigure}[b]{.48\linewidth}
            \includegraphics[width=\textwidth]{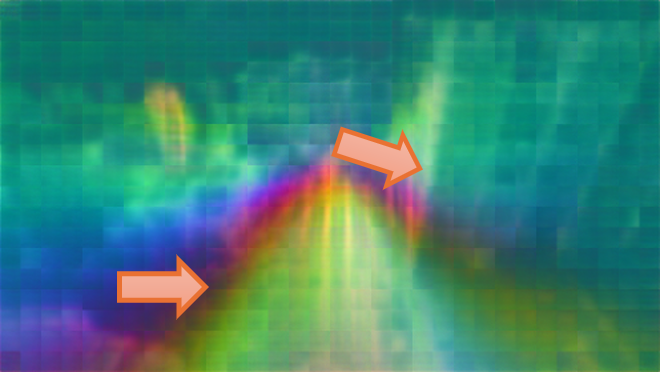}
            \caption{1st image descriptor,~\cite{vuong2025aerialmegadepth}}
            \label{dust3r_f1}
        \end{subfigure} &
        \begin{subfigure}[b]{.48\linewidth}
            \includegraphics[width=\textwidth]{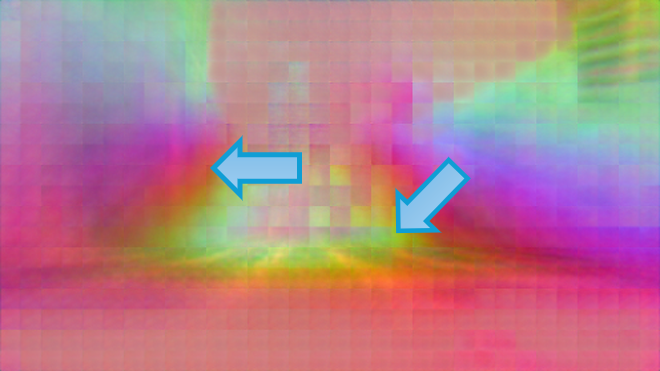}
            \caption{2nd image descriptor,~\cite{vuong2025aerialmegadepth}}
            \label{dust3r_f2}
        \end{subfigure} \\
        \begin{subfigure}[b]{.48\linewidth}
            \includegraphics[width=\textwidth]{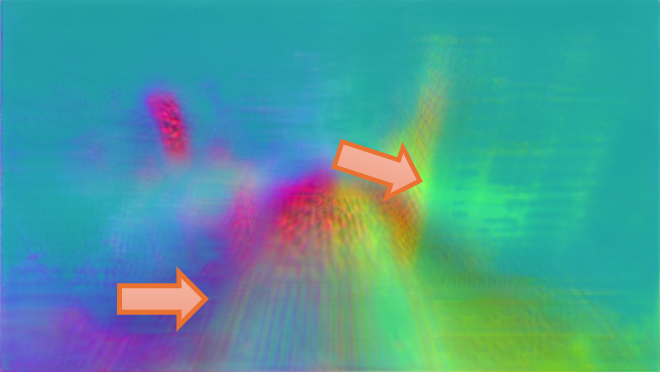}
            \caption{1st image descriptor, SPIDER}
            \label{ours_f1}
        \end{subfigure} &
        \begin{subfigure}[b]{.48\linewidth}
            \includegraphics[width=\textwidth]{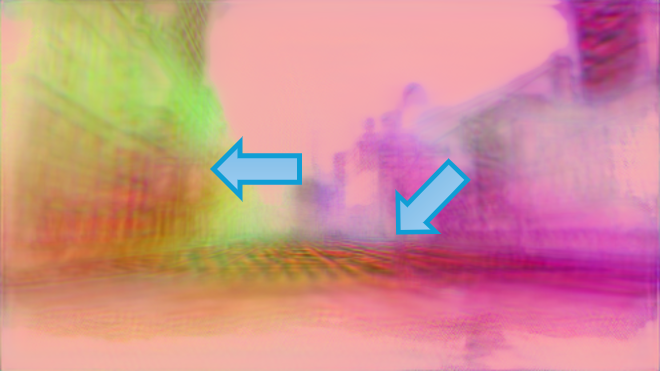}
            \caption{2nd image descriptor, SPIDER}
            \label{ours_f2}
        \end{subfigure} \\
    \end{tabular}
    \caption{Visualization of feature descriptors from Aerial-MASt3R %\cite{vuong2025aerialmegadepth}
    and SPIDER, based on images from the Teaser figure%~\cref{fig:teaser}. 
    By introducing a multi-scale feature upsampling, SPIDER obtains significantly better resolved features to achieve accurate correspondences.}
    \label{fig:feat}
\end{figure}

As shown in~\cref{ours_f1} and~\cref{ours_f2}, by progressively integrating geometric detail through the Fusion Gates, 
the Multi-Scale Descriptor Head 
yields descriptors that are significantly better resolved than previous approaches.

For training, we adopt a symmetric InfoNCE-based objective $\ell_{\text{match}}(i,j)$~\cite{leroy2024grounding} to supervise the descriptor head:
\begin{equation}
\ell_{\text{match}}(i,j) =
\log \frac{s_\tau(i,j)}{\sum_{k\in\mathcal{S}^A} s_\tau(k,j)} 
+ 
\log \frac{s_\tau(i,j)}{\sum_{k\in\mathcal{S}^B} s_\tau(i,k)},
\label{eq:match_loss}
\end{equation}
where $\{(i,j)\}$ are the groundtruth match indices in $\{I^A,I^B\}$ and the descriptor similarity is computed as
\begin{equation}
s_\tau(i,j) = \exp\!\left[-\tau \hat{D}_i^{A\top} \hat{D}_j^{B}\right],
\label{eq:similarity}
\end{equation}
with temperature parameter~$\tau$. 
Here, $\mathcal{S}^A$ and 
$\mathcal{S}^B$ 
denote valid pixel subsets in each image.
The final 3D training objective as
\begin{equation}
\mathcal{L}_{\text{feat}} =  
\sum_{(i,j)\in\hat{\mathcal{M}}} 
\hat{C}_{i,j}\, \ell_{\text{match}}(i,j)
- \alpha \log \hat{C}_{i,j},
\label{eq:conf_loss}
\end{equation}
where the mean confidence of a correspondence is
\begin{equation}
\hat{C}_{i,j} = (\hat{C}_i^A + \hat{C}_j^B) / 2.
\end{equation}

% Fast nearest-neighbor (FastNN) matching~\cite{leroy2024grounding} are employed to find correspondences.

\paragraph{Warp Head.} 
Geometry-driven matching focuses on high confidence regions where depth is relatively uniform. This can limit the match diversity in several ways, including when all the matches lie on one plane and lead to homography; furthermore, matching across wide baseline can be limited when depth is uncertain from one or both views.

Inspired by RoMa~\cite{edstedt2024roma}, we integrate a dense warp head with 3D VFM features, which focuses on matching 2D patterns. A transformer-based Match Decoder $\mathcal{D}$ first predicts an initial coarse warp field and confidence map at scale~16 from $F_\text{coarse}^A, F_\text{coarse}^B$:
% {\color{red} Variables are not defined here, also why hat? Don't use words to define networks, use variables}
\begin{equation}
\hat{W}_{16}^{A\rightarrow B}, \hat{C}_{16}^{A\rightarrow B} = \mathcal{D}(F_\text{coarse}^A, F_\text{coarse}^B),
\end{equation}
where $\hat{W}_{16}^{A\rightarrow B}, \hat{C}_{16}^{A\rightarrow B}$ represent the pixel-wise correspondences and confidence from $I^A$ to $I^B$.

The initial warp is subsequently refined in a hierarchical coarse-to-fine manner through a sequence of convolutional refinement modules~\cite{edstedt2024roma}:
\begin{equation}
\begin{split}
\hat{W}_{\frac{s}{2}}^{A\rightarrow B},\; \hat{C}_{\frac{s}{2}}^{A\rightarrow B} 
&= \mathop{\text{WarpRefiner}}\limits_{s\rightarrow \frac{s}{2}}\big(
    \hat{W}_{s}^{A\rightarrow B},\;
    \hat{C}_{s}^{A\rightarrow B}, \\
& \quad F^A_{\frac{s}{2}},\;
    F^B_{\frac{s}{2}}\big), \quad
    s\!\in\!\{16,8,4,2\}.
\end{split}
\label{eq:warp}
\end{equation}

% {\color{red} Is $F_1$ every used here?} 
After successive refinements, the final warp $\hat{W}_{1}^{A\rightarrow B}$ produces dense pixel-level correspondences aligned to the resized image and we can sample a certain number of matches from the warp based on the confidence map. 

For the coarse scale ($s{=}16$), the objective combines a cross-entropy term for anchor classification and a binary cross-entropy (BCE) term for confidence estimation:
\begin{equation}
\begin{split}
\mathcal{L}_{16} &=
\text{CE}\!\left(k^\dagger(x^B), k^\dagger(\hat{x}^B)\right)
 + \text{BCE}\!\left(C_{16}^{A\rightarrow B}, \hat{C}_{16}^{A\rightarrow B}\right).
\end{split}
\label{eq:coarse_loss}
\end{equation}
For finer scales, the refinement output at each scale~$s$  is formulated as a generalized Charbonnier~\cite{barron2019general} distribution 
(with $\alpha{=}0.5$) to penalize residual warp errors, together with a BCE term for confidence prediction:
\begin{equation}
\begin{split}
\mathcal{L}_{s} &=
\left(\|\hat{W}_{s}^{A\rightarrow B} - W_{s}^{A\rightarrow B}\|^2 + cs\right)^{\tfrac{1}{4}} \\
&\quad + \text{BCE}\!\left(C_{s}^{A\rightarrow B}, \hat{C}_{s}^{A\rightarrow B}\right),
\quad s \in \{8,4,2,1\}.
\end{split}
\label{eq:fine_loss}
\end{equation}

\paragraph{Two-Head Fusion.} We have explored several approaches to combine matches from the descriptor and warp head. Initial attempts are made at ensembling on a feature level, i.e., combining or replacing $F$ in \cref{eq:warp} with $\hat{P}$ from \cref{eq:descriptor} and using only the warp head matches, or vice versa. However, the objective function for one head tends to erase the pattern/geometry property from another head. We have also explored guidance-based match fusion, e.g., allowing the matches from one head to define regions to match for the other hand. This also limits the diversity of combined matches. In the end, we find evenly sampling the two heads based on confidence leads to the best performance.

\section{Experiments}
% We detail in \cref{s4.1} the training procedure of
% SPIDER. Then, we evaluate on several existing image matching benchmarks in \cref{s4.2}. Then we construct a new zero-shot benchmark for more challenging scenarios, with smaller overlap and visual ambiguities between images. We evaluate SPIDER and compare with other SoTA matchers in \cref{s4.3}. We finally validate our design choice with ablation studies in \cref{s4.ablation}.

\begin{table*}[]
\centering
\caption{SoTA comparison on existing image matching benchmarks, measured by AUC@5\degree. G.R + A.M denotes the concatenation of GIM-RoMa and Aerial-MASt3R. The best result is marked with \colorbox{best}{best} and second best with \colorbox{second}{second}.}
\resizebox{\linewidth}{!}{
\begin{tabular}{cl|c|ccc|cccccccccccc}
\toprule
 \multirow{2}*{Res.}      &  \multirow{2}*{Method} &  \multirow{2}*{\textbf{Mean ($\uparrow$)} } &    \multicolumn{3}{c|}{In-domain AUC$@5\degree\uparrow$} & \multicolumn{12}{c}{ZEB AUC$@5\degree\uparrow$}  \\
 & & & MegaDepth & ScanNet & AerialMega & GL3D & BLE    & ETI  & ETO  & KIT  & WEA   & SEA  & NIG  & MUL & SCE & ICL                  & GTA  \\
\midrule
% \multirow{5}{*}{low}  &  
% GIM-RoMa   &\cellcolor{second}46.7  & \cellcolor{best}48.9  & 27.1    & 40.1
% & \cellcolor{best}55.4 & 46.8   & \cellcolor{best}70.0 & \cellcolor{best}77.1 & 38.3       & 34.6                 & 51.4                 & 24.6   & 57.1                 & 29.0                 & 39.0       & 61.3                            \\
\multirow{5}{*}{low}  & MASt3R   &\cellcolor{second}45.5  & \cellcolor{second}40.0                  & 33.7    & 32.8 
& \cellcolor{second}52.6 & \cellcolor{best}48.3         & \cellcolor{second}49.2    & \cellcolor{second}55.7 & \cellcolor{second}43.0  & \cellcolor{second}34.9       & \cellcolor{best}51.9                & \cellcolor{second}25.3           & \cellcolor{second}63.0           & \cellcolor{second}34.5              & \cellcolor{best}56.4                   & \cellcolor{second}61.6                         \\
& Aerial-MASt3R  &43.7  & 40.0           & \cellcolor{second}34.1        & \cellcolor{second}49.3
& 47.3 & 41.7   & 45.3 & 51.6 & 40.8                 & 31.6                 & 50.4                 & 24.7                 & 60.0                 & 25.8                & 51.6                 & 61.0                          \\
& VGGT & 24.7& 38.4           & 33.9         & 21.9
& 11.2 & 10.3   & 45.8 & 55.6 & 35.3                 & 5.7                  & 2.3                  & 5.3                  & 48.3                 & 15.2                 & 32.9                 & 8.0                            \\  
& \textbf{SPIDER (Ours)} & \cellcolor{best}50.1  & \cellcolor{best}46.4 & \cellcolor{best}34.2   & \cellcolor{best}50.0     & \cellcolor{best}54.1  & \cellcolor{second}48.2   & \cellcolor{best}59.6 & \cellcolor{best}70.7   & \cellcolor{best}44.0    & \cellcolor{best}35.0     & \cellcolor{second}51.6 & \cellcolor{best}27.3     & \cellcolor{best}64.7    & \cellcolor{best}43.7     & \cellcolor{second}54.0 &\cellcolor{best} 65.7           \\ \midrule
\multirow{6}{*}{high} & 
RoMa & 48.5 & \cellcolor{best}{62.6} & 31.8  & 47.1 
& 48.3 &	40.6 &	\cellcolor{second}73.6	&79.8&	39.9	&34.4&	51.4	&24.2	&59.9	&33.7	&41.3	&59.2 
\\

& GIM-RoMa  & 51.4 & 56.3           & 28.6         & 47.4   
& 61.8 & 53.8   & \cellcolor{best}76.7 & \cellcolor{best}82.7 & 43.2                 & 36.7            & \cellcolor{best}53.2                 & 26.6                 & 60.7    & 33.8            & 45.4                 & 64.3                             \\
& MASt3R  & \cellcolor{second}54.1 & 52.6           & \cellcolor{second}32.0         & 43.8            
&\cellcolor{second}65.8 & \cellcolor{best}62.3 & 68.0 & 79.3 & \cellcolor{best}57.0      & \cellcolor{best}37.8                 & 52.8                 & 27.5                 & 62.6                 & \cellcolor{second}42.3                 & \cellcolor{second}57.1     &\cellcolor{second}70.0                          \\
& Aerial-MASt3R  &  51.0  & 53.2           & 29.9         & 49.7                               & 60.6 & 56.2   & 66.9 & 77.0 & 52.9                 & 35.2                 & 51.8                 & 25.5                 & 59.0                 & 30.1                 & 49.8                 & 66.6                               \\
& G.R. + A.M. & 52.9  & 57.2           & 29.9         & \cellcolor{second}50.2
& 63.4 & 57.6   & 68.3 & 74.8 & 46.3                 & 37.2                 & \cellcolor{second}53.0                 & \cellcolor{second}27.7                 & \cellcolor{second}64.0                 & 42.0                 & 53.0                 & 68.6                         \\ 
& \textbf{SPIDER (Ours)}   & \cellcolor{best}56.7 & \cellcolor{second}{60.0}           & \cellcolor{best}{33.4}         & \cellcolor{best}{58.7}                                                             & \cellcolor{best}66.4 & \cellcolor{second}59.8   & 69.1 & \cellcolor{second}82.3 & \cellcolor{second}54.7                 &\cellcolor{second} 37.5                 & 52.7                 & \cellcolor{best}28.1                 & \cellcolor{best}69.7                 & \cellcolor{best}48.6                 & \cellcolor{best}59.4                 & \cellcolor{best}71.5                              \\ \bottomrule
\end{tabular}
}
\vspace{-0.5em}
\label{tab:zeb}
\end{table*}

% \subsection{Training}\label{s4.1}
\textbf{Training Data.}\label{s4.0} We train our framework on ten large-scale datasets, covering synthetic and real-world, indoor and outdoor, and aerial-to-ground scenes. The training corpus includes Habitat~\cite{savva2019habitat}, ARKitScenes~\cite{baruch2021arkitscenes}, BlendedMVS~\cite{yao2020blendedmvs}, MegaDepth~\cite{li2018megadepth}, Static Scenes 3D~\cite{schroppel2022benchmark}, ScanNet++~\cite{yeshwanth2023scannet++}, CO3D-v2~\cite{reizenstein2021common}, Waymo~\cite{sun2020scalability}, WildRGB~\cite{xia2024rgbd}, and Aerial-MegaDepth~\cite{vuong2025aerialmegadepth}. Together, these datasets provide over 10 million image pairs. All images are resized such that their maximum dimension is 512 pixels, while preserving aspect ratio to maintain geometric consistency.

\noindent\textbf{Training Configuration.}
We adopt a ViT-Large~\cite{dosovitskiy2020vit} backbone as the image encoder and a ViT-Base~\cite{dosovitskiy2020vit} architecture for the decoders, initialized with pretrained weights from Aerial-MASt3R~\cite{vuong2025aerialmegadepth}. The ViT backbone is kept frozen during training.
While SPIDER is conceptually designed with dual correspondence heads to balance sensitivity and specificity, we find in practice that joint training of the two heads is suboptimal (see \cref{s4.ablation}). Instead, we train each head independently with its dedicated fine encoder, which both are initialized from pretrained VGG-19. 
% At inference time, for each image pair, we need run it with the image order of ($I^A, I^B$) and ($I^B,I^A$).

The dense warp head is trained for 160 epochs, sampling 40K image pairs per epoch, while the feature head is trained for 50 epochs with 400K pairs per epoch. Pairs are equally distributed between all datasets. We use the AdamW optimizer with an initial learning rate of $1\times10^{-4}$, applying linear warmup followed by cosine decay. All experiments are conducted on four NVIDIA H200 GPUs. See supplementary material for more details.
\begin{figure}[!htb]
    \setlength{\abovecaptionskip}{3pt}
    \setlength{\tabcolsep}{2pt}
    \begin{tabular}[b]{cc}
        \begin{subfigure}[b]{.48\linewidth}
            \includegraphics[width=\textwidth]{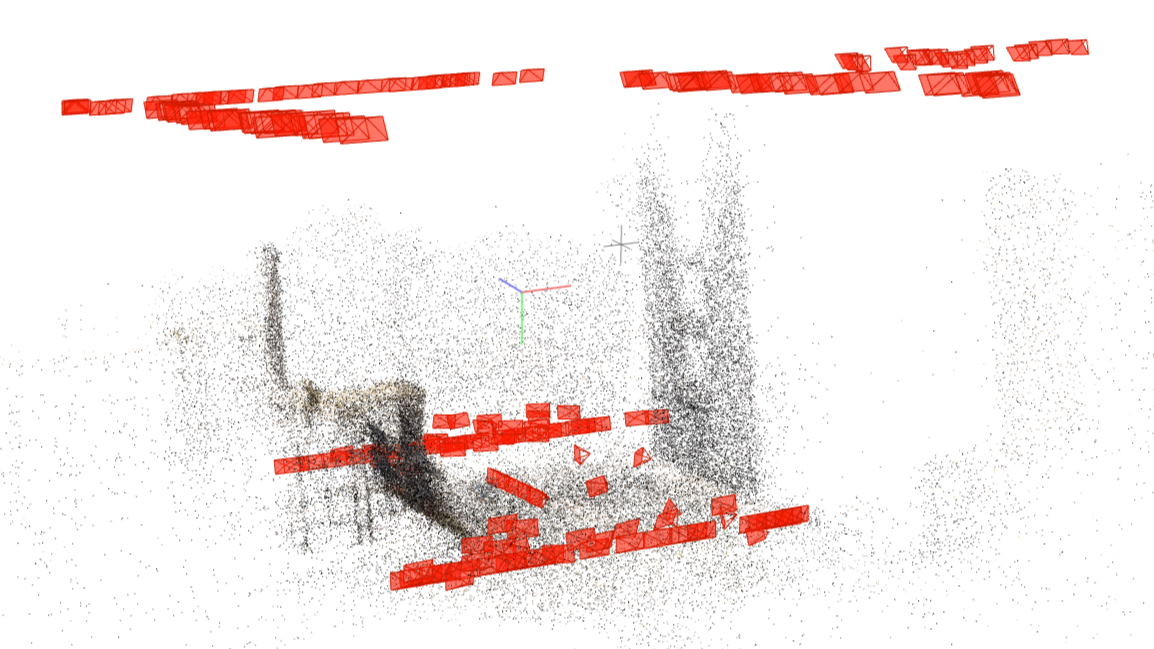}
            \caption{Urban}
            \label{s01}
        \end{subfigure} &
        \begin{subfigure}[b]{.48\linewidth}
            \includegraphics[width=\textwidth]{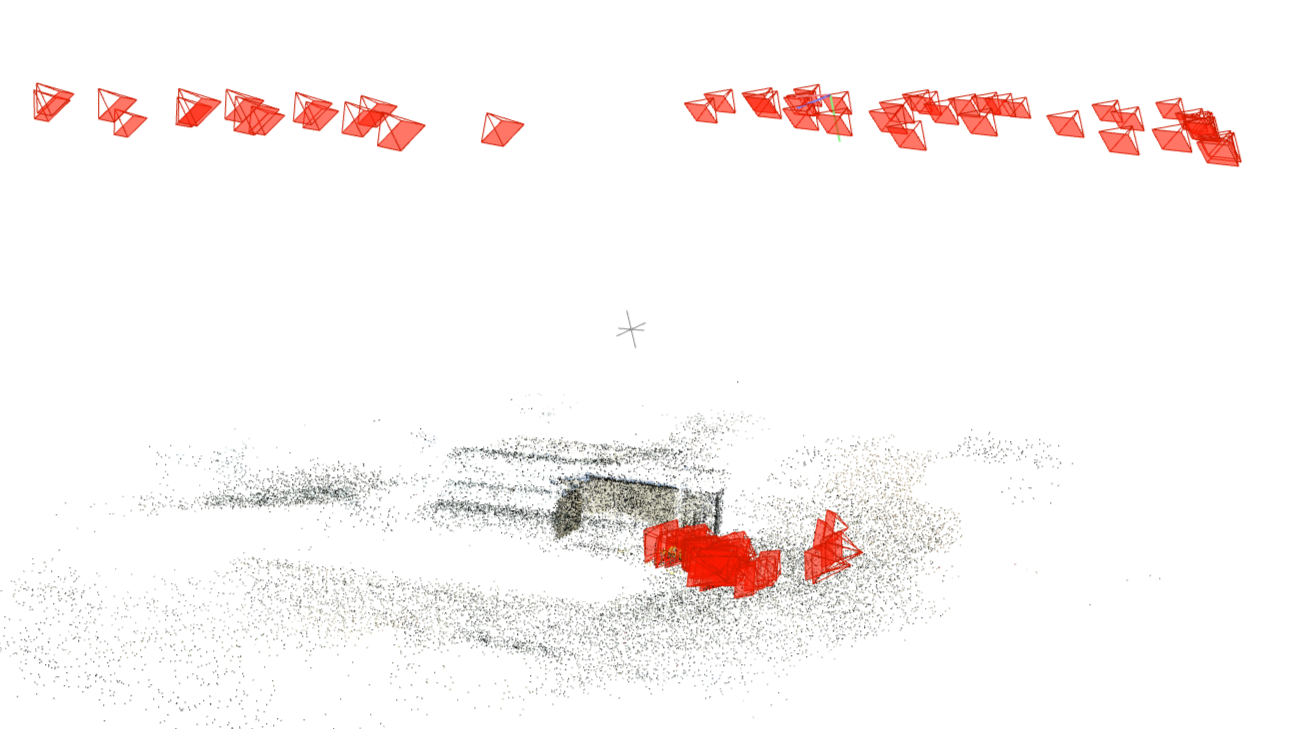}
            \caption{Warehouse}
            \label{m07}
        \end{subfigure} \\
        \begin{subfigure}[b]{.48\linewidth}
            \includegraphics[width=\textwidth]{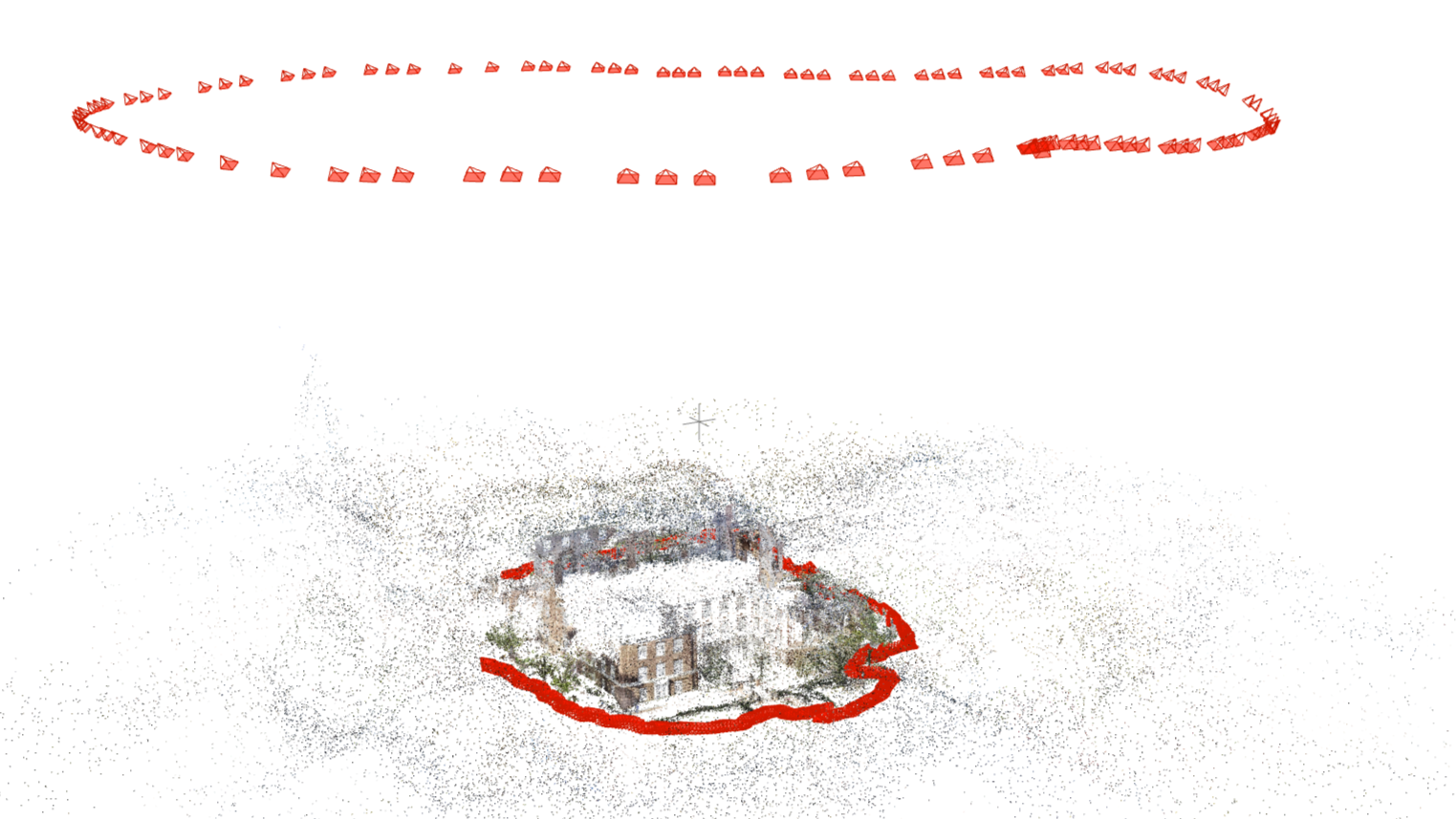}
            \caption{Campus 1}
            \label{ames}
        \end{subfigure} &
        \begin{subfigure}[b]{.48\linewidth}
            \includegraphics[width=\textwidth]{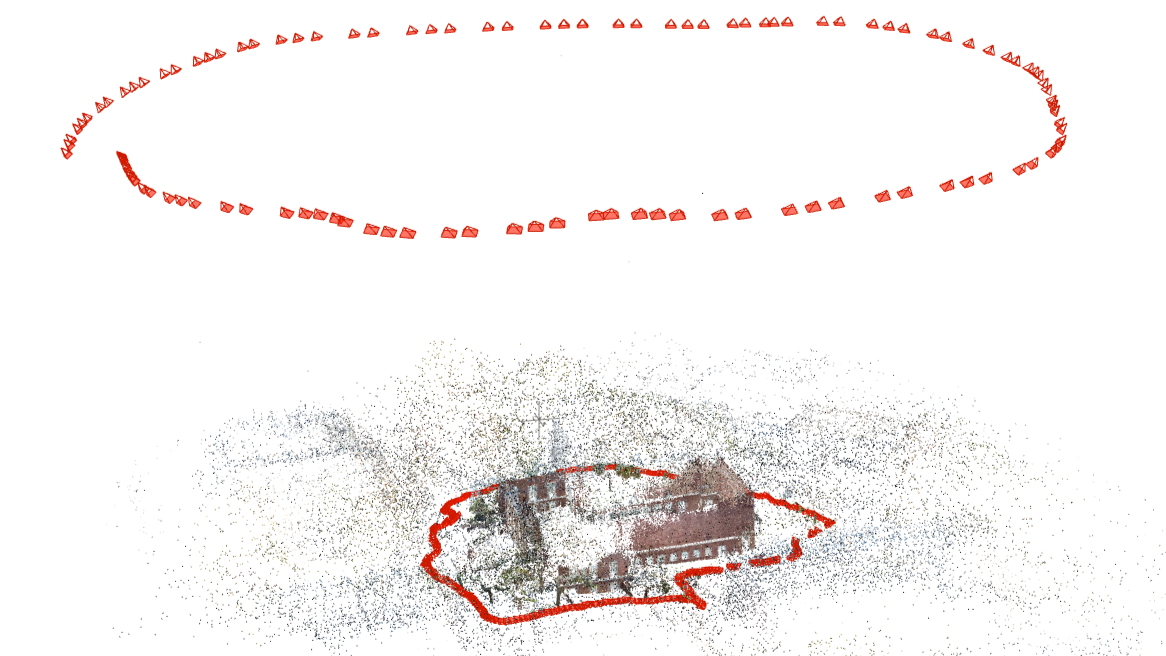}
            \caption{Campus 2}
            \label{shriver}
        \end{subfigure} \\
    \end{tabular}
    \caption{Visualization of camera positions for four multi-elevation scenes collected in unconstrained scenarios.}
    \vspace{-1em}
    \label{fig:colmap}
\end{figure}

\subsection{Two-View Geometry Evaluation} \label{s4.2}
% Due to differences in training and matching strategies across existing methods, 
We evaluate our approach against current SoTA on a diverse set of two-view geometry benchmarks. Since some methods natively match in low ($\sim$512px)~\cite{leroy2024grounding,vuong2025aerialmegadepth,wang2025vggt} or high ($\sim$1600px)~\cite{edstedt2024roma,shen2024gim} resolutions, we test under both image settings. Results are summarized in Table~\ref{tab:zeb}. 

\noindent\textbf{In-Domain Benchmarks.} We evaluate our method on three two-view geometry datasets covering outdoor, indoor, and aerial-to-ground scenarios, from which SPIDER uses the training sets from. 
MegaDepth-1500~\cite{li2018megadepth} and ScanNet-1500~\cite{yeshwanth2023scannet++} serve as standard outdoor and indoor benchmarks, and we follow the evaluation protocols of~\cite{sarlin2020superglue,chen2022aspanformer,sun2021loftr,edstedt2023dkm,edstedt2024roma}. 
Since there is currently no public benchmarks for aerial-to-ground matching, we construct the Aerial-MegaDepth~\cite{vuong2025aerialmegadepth} subset by selecting the same scene used in MegaDepth-1500~\cite{li2018megadepth}, i.e. \textit{Scene 15 (St. Peter’s Basilica)} and \textit{Scene 22 (Brandenburger Tor)}, with covisibility ratios between 0.1 and 0.2. The subset contains 3,355 image pairs in total. From models that are not trained on~\cite{vuong2025aerialmegadepth}, RoMA~\cite{edstedt2024roma}/GIM-RoMA~\cite{shen2024gim} achieves significantly better performances, showing that patterns are easier to match compared to geometry in extreme conditions. However, MASt3R~\cite{leroy2024grounding} performs the best in indoor scenarios where textureless regions appear more often. Our method achieves new SoTA performance by integrating both pattern and geometry matching methods, with notably large gains on aerial-to-ground scenarios. In comparison, \emph{simply concatenating matches} from~\cite{shen2024gim} and Aerial-MASt3R~\cite{vuong2025aerialmegadepth} leads to worse performance than SPIDER. In supplemental material, we provide a breakdown of the performances from SPIDER's individual heads. 
 
 % ScanNet exhibits a slight drop from low to high resolution due to the object-centric nature of its test scenes. In contrast, both MegaDepth and Aerial-MegaDepth benefit from higher input resolution, as their images contain richer high-frequency details and cover larger real-world scales.

\noindent\textbf{Zero-Shot Benchmarks.}
We further test generalization under a zero-shot setting, where all test datasets are unseen during training. ZEB~\cite{shen2024gim} includes 12 datasets spanning a wide variety of scenes, serving as a strong benchmark for generalized image matching evaluation. RoMa~\cite{edstedt2024roma}, which is only trained and validated on MegaDepth~\cite{li2018megadepth}, generalizes worse on ZEB~\cite{shen2024gim}. We note that MASt3R~\cite{leroy2024grounding} has been trained on \underline{BLE} and \underline{KIT} within the ZEB. Our method achieves strong zero-shot performance, achieving \textbf{+5.7}, \textbf{+6.3}, \textbf{+2.3} in MUL, SCE, ICL compared to the next best, and an overall \textbf{+5.7} compared to Aerial-MAST3R~\cite{vuong2025aerialmegadepth}, which SPIDER uses as for its backbone. 
% confirming its ability to handle drastic appearance and geometric variations without retraining. 
For comparisons of more baselines in ZEB~\cite{shen2024gim}, a more comprehensive study is done in~\cite{shen2024gim}. 

% \begin{figure}
%     \centering
%     \includegraphics[width=1\linewidth]{fig/ames.jpg}
%     \caption{Example of aerial-to-ground pairs}
%     \label{fig:ames}
% \end{figure}

\subsection{Unconstrained Evaluation}\label{s4.3}
% Existing zero-shot benchmarks such as ZEB provide an important step toward evaluating correspondence methods under distribution shift; however, they remain limited in their ability to reflect truly unconstrained real-world scenarios. 
Although ZEB merges twelve heterogeneous datasets, its image pairs still fall within moderate difficulty: overlap ratios are typically between 10–50\%, viewpoints remain largely co-planar, and without aerial-to-ground pairs. 
% This leads to a benchmark that is diverse but still biased toward ground-level imagery, where field-of-view, scene scale, and geometry are relatively consistent, avoiding t
Extreme viewpoint, scale, and overlap conditions ($<$10\%) from unconstrained images are not considered. 
% Another limitation is that 
Conventional benchmarks also contain few visual ambiguities~\cite{cai2023doppelgangers}, where visually similar but geometrically distinct areas produce false correspondences, which are particularly prevalent in buildings. 

\noindent\textbf{Datasets.} To address these gaps, we collected images from four distinct scenes to evaluate matching robustness in unconstrained environments: 
% UBench explicitly includes aerial-to-ground pairs, capturing the extreme geometric and visual variations that arise in real zero-shot deployment.
% We select four representative scenes for the proposed benchmark: 
one synthetic city scene with multiple symmetric buildings (\emph{Urban}), one real scene from the ULTRRA Challenge~\cite{2zs6-ht63-24} on a cluster of warehouses (\emph{Warehouses}), and two real scenes on a single campus building (\emph{Campus 1 and 2}). All scenes contain complex building structures; while only \emph{Urban} has perfect pose groundtruth, all real scenes are calibrated using COLMAP constrained by RTK-grade GPS.
% Images are paired based on latitude–longitude proximity to promote meaningful yet challenging cross-view relationships, and all candidate pairs are further manually inspected to ensure the presence of visual ambiguity and difficult geometric configurations. 
In total, the benchmark contains 3000 image pairs. \emph{Urban} images has larger Field-of-View than the other scenes, but also exhibits the strongest visual ambiguities, as ground images are captured on the same street from opposite viewing directions.
% (see \cref{s01} and the example in \cref{fig:teaser}). 
\emph{Warehouses} has a small coverage of ground and larger aerial coverage, where the aerial images are captured from opposite directions. \emph{Campus 1 and 2} contain aerial cameras that are much higher from the ground, making the image scales drastically different. We extract 2400 aerial-to-ground image pairs out of these four scenes to comprehensively analyze the robustness of current methods. Given the complexity of the \emph{Urban} scene and its perfect groundtruth, we also extract 600 ground-to-ground pairs with potential visual ambiguities. 

\noindent\textbf{Metrics.} As shown in \cref{tab:uben}, we evaluate using unconstrained matching with camera pose error AUC@20\degree. SPIDER achieves the highest average performance. VGGT~\cite{wang2025vggt} and MASt3R~\cite{leroy2024grounding} failed on this aerial-to-ground benchmark. Again, RoMa~\cite{edstedt2024roma} demonstrates impressive accuracy in \emph{Campus 1 and 2} aerial-to-ground matching without training on~\cite{vuong2025aerialmegadepth}. 
However, in \emph{Urban}'s ground-to-ground matching, MASt3R~\cite{leroy2024grounding} is much better than RoMa~\cite{edstedt2024roma}, demonstrating robustness to visual ambiguities in geometry-based matching. Even in aerial-to-ground scenarios, RoMa~\cite{edstedt2024roma} fails in multi-building scenarios in \emph{Urban and Warehouses} due to visual ambiguities from similar buildings. VGGT~\cite{wang2025vggt} exhibits severe out-of-domain failures, likely because its feedforward camera pose prediction is domain-specific and is not generalizable enough compared to COLMAP~\cite{schoenberger2016sfm}.

% Scene 1 is highly ambiguous, we found more ground-to-ground pairs to be visual ambiguous. See supplementary. Scene 2 small coverage of ground and larger aerial coverage, post a harder scenarios. Scene 3 and Scene 4

\begin{table}[t]
\centering
\setlength{\tabcolsep}{0.1em}
\footnotesize
\caption{SoTA comparison on two-view geometry. For Urban scene, both Aerial-Ground (A+G) and Ground-Ground (G+G) pairs are evaluated.}
\label{tab:uben}
\begin{tabular}{lcccccc}
\toprule
 & Urban & Urban-G & Warehouses & Campus 1 & Campus 2 & Mean \\
 Cam. Position & A+G & G+G & A+G & A+G & A+G & \\
\midrule
RoMa & 5.9&12.3 &2.4  & 21.5 & 18.0 & 12.0 \\
GIM-RoMa& 28.7  &15.3& 3.4 & 11.6&2.6 & 12.3 \\
VGGT & 11.0 & 10.9&3.0 & 5.1 & 0.9 &  6.2\\
MASt3R & 10.2 & 26.8&0.4 & 0.0 & 0.0 &  7.5 \\
Aerial-MASt3R& \cellcolor{second}60.5 & \cellcolor{second}36.7&\cellcolor{second}18.6  & \cellcolor{second}33.0 & \cellcolor{second}26.9 & \cellcolor{second}35.1\\
G.R + A.M &51.8&30.1 & 9.7 & 31.2 & 21.5 & 28.9 \\
SPIDER& \cellcolor{best}70.8& \cellcolor{best}56.7 & \cellcolor{best}20.9 & \cellcolor{best}33.3& \cellcolor{best}28.3& \cellcolor{best}42.0\\
\bottomrule
\end{tabular}
\end{table}

% \input{fig/example_vis}
% \begin{figure*}
%     \centering
%     \includegraphics[width=1\linewidth]{fig/example_vis.png}
%     \caption{Enter Caption}
%     \label{fig:placeholder}
% \end{figure*}

% \begin{figure}
%     \centering
%     \includegraphics[width=0.75\linewidth]{fig/auc_shared_vs_separate.pdf}
%     \caption{\textbf{Ablation on shared vs. separate ConvNets.} 
% Comparison of joint and separate training for warp and feature heads. 
% Sharing ConvNets degrades warp performance, while separate ConvNets yield higher overall accuracy.}
%    \label{tab:cnn_ablation}
%    \vspace{-10pt}
% \end{figure}

% \begin{figure*}
%     \centering
%     \includegraphics[width=1\linewidth]{fig/planarity.jpg}
%     \caption{Qualitative visualization of planarity bias. Each column shows dense correspondences between two views. IR denotes the inlier ratio from homography estimation—higher values indicate that most correspondences lie on dominant planar regions.}
%     \label{fig:planarity_vis}
% \end{figure*}
\subsection{Visualization and Discussion}\label{s4.ablation}
\begin{figure*}
    \centering
    \includegraphics[width=1\linewidth]{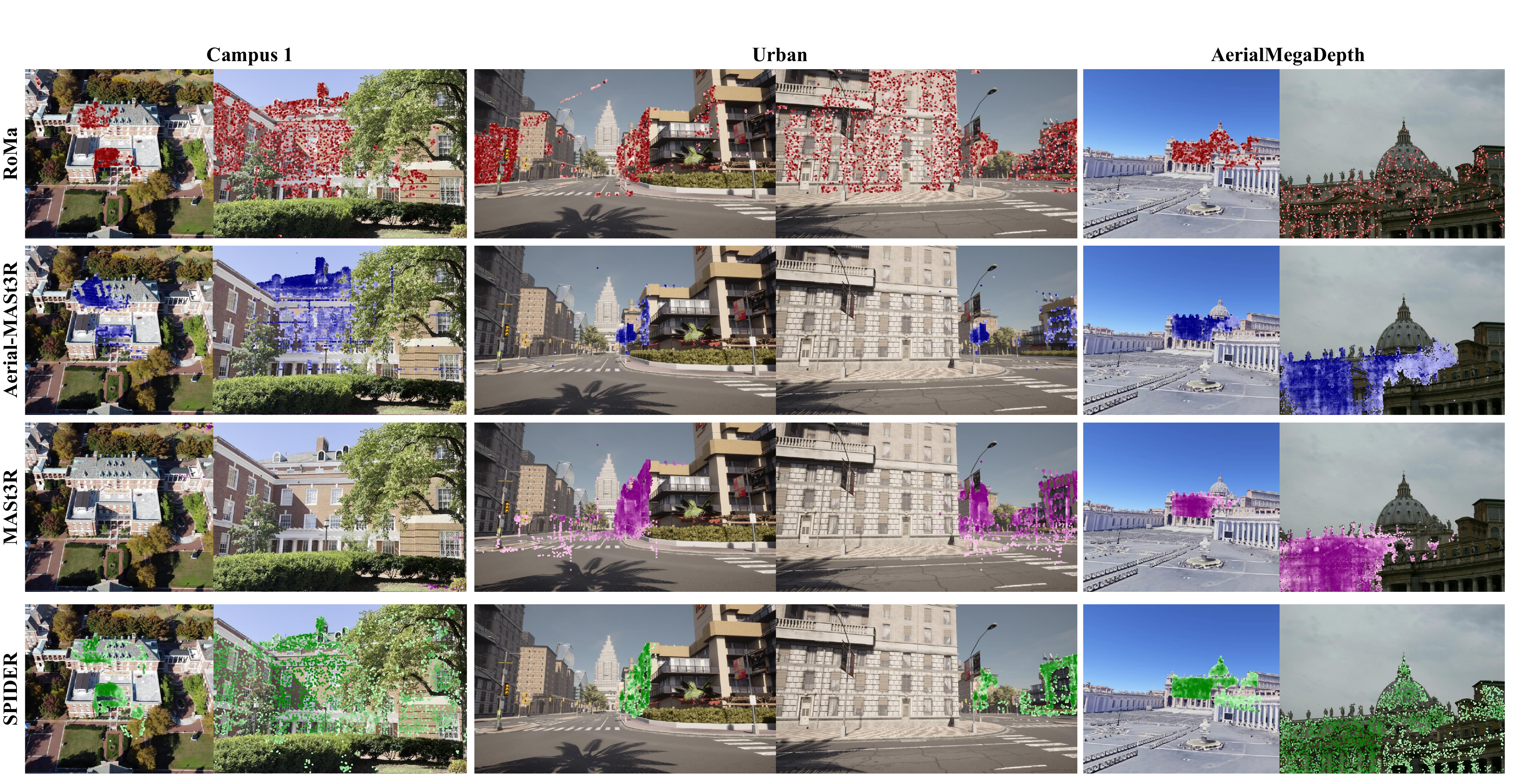}
    \caption{\textbf{Visual Comparison under unconstrained settings.} Image pattern-driven methods, e.g. RoMa~\cite{rombach2022high}, finds diverse matches across many planes; however, matches may be false negatives on two sides of the building. Geometry-driven methods~\cite{leroy2024grounding,vuong2025aerialmegadepth} are better at matching planes. This can lead to homography if a  confident plane dominates, e.g., when Aerial-MASt3r matches the wrong signs in \emph{Urban} with high confidence. SPIDER combines both approaches and produces diverse and accurate matches.}
    \label{fig:vis}
\end{figure*}

By training on~\cite{vuong2025aerialmegadepth}, Aerial-MASt3R is much better than other methods in~\cref{tab:uben}; however, as shown in~\cref{fig:vis}, it remains less robust in certain scenarios.

% In this section, we analyze the effectiveness of our design choices from the following aspects: (1) shared vs.\ separate fine encoders for the dual heads, (2) learnable vs.\ frozen backbone, and (3) the effect of multi-scale refinement on mitigating planarity bias. 
% All models are trained on the full set of 10 datasets introduced in \cref{s4.0}, and evaluated on the MegaDepth test split unless otherwise specified.

\noindent\textbf{Effect of Multi-Scale 3D Feature Refinement.}
Although Aerial-MASt3R also includes a 3D feature head, its descriptors are directly regressed from the coarse ViT features at scale~16 through a MLP, without incorporating finer-scale geometric cues. 
As a result, the predicted descriptors mainly encode global depth trends but lack local geometric discrimination--causing many matches to collapse onto dominant planes such as the ground or façades. 
In contrast, our 3D head progressively refines features from multiple scales ($\hat{P}_{16}\!\rightarrow\!\hat{P}_8\!\rightarrow\!\hat{P}_4\!\rightarrow\!\hat{P}_2\!\rightarrow\!\hat{P}_1$) using attention-based FusionGates, which inject local spatial structure into the representation. 
This hierarchical refinement allows descriptors to preserve cross-view geometric consistency while maintaining local details, effectively reducing planar bias. 
Quantitatively, the planarity ratio drops \textbf{$\sim$30\%} from 0.22 in Aerial-MASt3R to 0.16 in our feature head (\cref{tab:planarity_ratio}), and the correspondences become more spatially diverse across different depth layers. 

This can be visualized both in a AerialMegaDepth scene and the \emph{Urban} scene. Aerial-MASt3R produces high confidence matches on regions of uniform depth, e.g., the front building in AerialMegaDepth and the sign in \emph{Urban} (see \cref{fig:vis}), which dominate geometric verification and lead to the removal of other matches that are less confident. Similar depth can also lead to ambiguities, as is the case of the matching signs in \emph{Urban} that do not belong to the same object. In comparison, SPIDER is able to propose \emph{confident and diverse} matches across the scenes, leading to better performance overall.

\begin{table}[t]
\centering
\caption{\textbf{Planarity Ratio comparison on Aerial-MegaDepth.} 
Planarity Ratio is defined as the proportion of image pairs with homography inlier ratio $>$ 0.5 among those with fundamental matrix inlier ratio $>$ 0.6. 
Lower values indicate less planar bias and greater geometric diversity.}
\label{tab:planarity_ratio}
\small
\setlength{\tabcolsep}{5pt}
\resizebox{\linewidth}{!}{
\begin{tabular}{@{}lccccc@{}}
\toprule
 & RoMa & GIM-RoMa & MASt3R & Aerial-MASt3R & \textbf{SPIDER} \\
\midrule
Planarity Ratio ($\downarrow$) & 0.20 & 0.25 & 0.27 & 0.22 & \textbf{0.16} \\
\bottomrule
\end{tabular}
}
\end{table}

\begin{table}[t]
\centering
\caption{\textbf{Ablation on backbone training strategy.} 
Comparison between fully learnable and frozen backbones on MegaDepth. 
Metrics are PCK@5 on the training set and AUC@5° on the test set. 
Freezing the backbone reduces both parameter count and training time while improving generalization.}
\label{tab:backbone_ablation}
\resizebox{\linewidth}{!}{
\begin{tabular}{lcccc}
\toprule
\textbf{Backbone} & \textbf{\#Params (M)} & \textbf{Time (h)} & \textbf{PCK@5$\uparrow$} & \textbf{AUC@5\degree $\uparrow$} \\
\midrule
Learnable & 642.4 & 108 & 96.1 & 52.3 \\
Frozen    & 111.7 & 65  & 95.1 & 57.6 \\
\bottomrule
\end{tabular}
}
\end{table}

\noindent\textbf{Ablation Study.} 
As discussed in Sec.~\ref{s3.1}, we choose to train lightweight output heads on top of the Aerial-MASt3R features rather than fine-tuning all model weights to better leverage the pretrained representations. We compare two strategies: a fully learnable backbone versus a frozen backbone with trainable heads only. As shown in Table~\ref{tab:backbone_ablation}, freezing the backbone reduces the total number of parameters from 642M to 112M and shortens training time by nearly 40\%. Despite a slightly lower PCK@5 on the training set, the frozen configuration achieves a significantly higher AUC@5° on the test set (+5.3), indicating stronger generalization. This result suggests that the pretrained geometric representations in the backbone already encode robust feature priors, and fine-tuning the entire network may lead to overfitting to dataset-specific scene statistics.
 
 We then analyze the fine encoders. Despite sharing a coarse-to-fine structure, the warp head regresses continuous coordinates whereas the feature head performs discrete discrimination, causing gradient conflicts under joint training. \cref{tab:cnn} reports results for shared and separate ConvNets. Specifically, the warp head performance drops notably. The reason is that we directly predict coordinates, and this process becomes unstable when they are influenced by the feature matching supervision.   For a comprehensive ablation study on the design of our Multi-Scale Descriptor Head, various approaches in ensembling two-head matches, etc., please refer to the supplemental material.
\begin{table}[t]
\centering
\small
\caption{\textbf{Ablation on design of fine encoders.} Comparison between shared and separate ConvNets for two matching heads Metrics are AUC@5\degree on MegaDepth at resolution of 512.}
\label{tab:cnn}
\begin{tabular}{lcccc}
\toprule
 ConvNets & Warp Head & Feature head & Combine \\
 \midrule
 Shared & 43.9 & 43.7 & 45.1 \\
 Separate &45.5 & 43.1 & 46.7 \\
\bottomrule
\end{tabular}
\end{table}

\noindent\textbf{Limitation.} Despite SPIDER's impressive performance, the dual-head architecture and multi-scale refinement introduce some compute overhead, leading to around thirty percent additional inference time compared to Aerial-MASt3R~\cite{vuong2025aerialmegadepth}. Second, while the ensemble of warp- and descriptor-based correspondences improves performance, 
the ensemble is not trivial and can lead to inconsistent matches, especially in highly repetitive structures or extremely low-overlap views where the two heads may disagree. Since the confidence value for each head is with respect to different objective functions, the best way to aggregate matches remain an open question.

\section{Conclusion}
We presented SPIDER, a dual-head coarse-to-fine matching framework that unifies semantic alignment and geometric consistency by leveraging 3D-pretrained foundation features.
Through a warp head and a feature head, SPIDER jointly predicts dense correspondence fields and geometry-aware descriptors, enabling robust image matching across large viewpoint and scale variations.
Extensive experiments demonstrate that SPIDER achieves state-of-the-art performance on both in-domain and zero-shot benchmarks, including highly challenging aerial-to-ground scenarios.
Our analysis further shows that multi-scale coarse-to-fine refinement effectively mitigates planarity bias, and rejects visual ambiguity. We believe this work provides a strong step toward bridging 2D and 3D representation learning, opening new directions for geometry-aware vision foundation models and unified camera calibration frameworks.
\section{Acknowledgement}
This research is based upon work supported by the Office of the Director of National Intelligence
(ODNI), Intelligence Advanced Research Projects Activity (IARPA), via IARPA R\&D Contract
No. 140D0423C0076. The views and conclusions contained herein are those of the authors and
should not be interpreted as necessarily representing the official policies or endorsements, either
expressed or implied, of the ODNI, IARPA, or the U.S. Government. The U.S. Government is
authorized to reproduce and distribute reprints for Governmental purposes notwithstanding any
copyright annotation thereon.
\clearpage
\setcounter{page}{1}
\setcounter{section}{0}
\maketitlesupplementary
\renewcommand\thesection{\Alph{section}}
\section{More Implementation Details}
We adopt a VGG-19 backbone for the fine encoder. Intermediate features are extracted from layers {40, 27, 14, 7}, which correspond to spatial scales {8, 4, 2, 1}, respectively. These features are integrated with coarse feature from 3D VFMs at scale 16 into a five-level feature pyramid at scales {16, 8, 4, 2, 1}. For the warp head, feature channels are projected to dimensions {512, 512, 256, 64, 9}, respectively. For the feature head, before projecting features into the final hidden space of dimension 128, we first map the pyramid features to intermediate dimensions {256, 128, 128, 64, 64}.

At inference time, for each image pair, we perform a bidirectional evaluation by running the network on both input orders: ($I^A, I^B$) and ($I^B,I^A$).

\section{Linear Probing 2D and 3D VFMs}
 To justify our use of 3D-pretrained representations, we conduct a linear-probe experiment comparing both 2D- and 3D-pretrained backbones on an image matching task. The evaluated backbones include:
VGG19, ResNet-50, Stable Diffusion, AM-RADIO, DINOv2, DINOv3, DUNE, DUSt3R, MASt3R, Aerial-DUSt3R, Aerial-MASt3R, and VGGT.  Following the protocol of~\cite{edstedt2024roma}, we freeze the pretrained encoder and/or decoder and train a single linear projection layer on top of it, while correspondences are then established via kernel nearest-neighbor matching.

 We evaluate performance on MegaDepth (outdoor scenes) and Aerial-MegaDepth (aerial-to-ground) test sets, using two standard metrics: (1) end-point-error (EPE), averaged at a standardized resolution of 448×448 to measure precision, and (2) PCK@32, the percentage of matches with reprojection error below 32 pixel to measure robustness.

 As shown in Table~\ref{tab:2dvs3d}, 3D-pretrained backbones outperform conventional 2D-pretrained ones. Even though the 3D-pretrained models are exposed to only around 10 million image pairs during training—orders of magnitude fewer than the 142 million and 1.7 billion single-view images used by DINOv2 and DINOv3—their encoder representations perform comparably well.
This indicates that multi-view supervision provides a far more informative learning signal than purely 2D appearance-based training. As observed by Chen et al.~\cite{chen2025easi3r}, the cross-attention mechanism in the decoder implicitly learns rigid view transformations and can extract motions from these layers. We find that a similar mechanism also benefits image matching. When equipped with a decoder, 3D-pretrained models outperform both their encoders and all 2D counterparts by a substantial margin. Among all candidates, Aerial-MASt3R decoded feature achieves the best overall performance, with the lowest EPE (8.3) and highest PCK@32 (96.6\%), while VGGT decoder ranks second.

\section{Baselines Evaluation Protocol}
All baseline methods are evaluated under the same protocol to ensure comparability. For each image pair, we first obtain image correspondences from the baseline method and then estimate the relative camera pose by computing the essential matrix using the ground-truth camera intrinsics. This procedure is applied consistently across all local and dense matching methods.

MASt3R and Aerial-MASt3R, in their official implementation, estimate camera pose via global alignment (GA) by aligning 2D--3D correspondences with its predicted pointmap. For completeness, we additionally report MASt3R’s GA-based pose results for in-domain benchmarks in \cref{tab:mast3r-ga}. Across all three datasets, the E-Matrix protocol yields substantially higher AUC@5\degree~scores for MASt3R, indicating that its native GA alignment underperforms when applied to large-baseline or noisy correspondences. Aerial-MASt3R also benefits on ScanNet and Aerial-MegaDepth, where the E-Matrix protocol improves robustness in scenes with strong viewpoint changes or complex geometry. Because GA is incompatible with the essential-matrix protocol used for all other baselines—and empirically produces inferior pose estimates—the main paper reports MASt3R and Aerial-MASt3R using the unified E-Matrix protocol for a fair comparison.

\begin{table}[t]
\centering
\caption{\textbf{Comparison of MASt3R and Aerial-MASt3R under different pose-estimation protocols.} 
We report AUC@5\degree~for three in-domain benchmarks. ``GA'' denotes the official global-alignment procedure, while ``E-Matrix'' denotes our unified essential-matrix protocol used for all baselines.}
\resizebox{\linewidth}{!}{
\begin{tabular}{lccc}
\toprule
\textbf{Method} & \textbf{MegaDepth} & \textbf{ScanNet} & \textbf{AerialMega} \\
\midrule
MASt3R (GA)          & 35.7 & 31.9 & 20.8 \\
MASt3R (E-Matrix)    & \textbf{40.0} & \textbf{33.7} & \textbf{32.8} \\
Aerial-MASt3R (GA)   & \textbf{42.8} & 29.3 & 38.7 \\
Aerial-MASt3R (E-Matrix) & 40.0 & \textbf{34.1} & \textbf{49.3} \\
\bottomrule
\end{tabular}
}
\label{tab:mast3r-ga}
\end{table}

VGGT differs from other baselines because it includes a dedicated camera-pose head that directly regresses the relative rotation and translation. For VGGT, we use its predicted pose without applying essential-matrix estimation. 

Baseline methods also differ in their training and inference resolutions. RoMa and GIM-RoMa are trained at a fixed resolution of $672\times672$ and can be upsampled to $1344\times1344$ during inference. DUSt3R, MASt3R, Aerial-MASt3R, VGGT, and our SPIDER are trained with a maximum image dimension of 512 pixels. 
We treat this $512$-pixel input size as the \emph{low-resolution} setting for all baselines. MASt3R additionally employs a coarse-to-fine cropping strategy that splits the image into overlapping 512-pixel crops and fuses their predictions to recover dense correspondences at the original resolution. We use whose predictions are fused into dense, full-resolution correspondences. We use this coarse-to-fine variant, capped at a maximum input dimension of 1600 pixels, as the \emph{high-resolution} setting. Finally, to avoid directional bias, all symmetric matching methods are run in both input orders $(I^A,I^B)$ and $(I^B,I^A)$, with predictions fused after geometric verification.

\section{More Ablations}

\begin{table}[t]
\centering
\caption{\textbf{Ablation on Multi-scale Descriptor design.} 
Comparison of different refinement variants on MegaDepth-1500 using AUC@5\degree, AUC@10\degree, and AUC@20\degree at \emph{low resolution}. 
The difference between ``MSF --$\alpha$'' and ``MSF +$\alpha$ (Ours)'' is the presence of the predicted gating coefficient $\alpha$.}
\resizebox{\linewidth}{!}{
\begin{tabular}{lccc}
\toprule
\textbf{Method} 
& \textbf{AUC@5\degree} 
& \textbf{AUC@10\degree} 
& \textbf{AUC@20\degree} \\
\midrule
\textit{Coarse-only}    & 42.0 & 60.1 & 75.0 \\
\textit{FPN-style}      & 39.9 & 57.8 & 73.0 \\
\textit{MSF --~$\alpha$} & 39.5 & 58.2 & 73.8 \\
\textit{\textbf{MSF +~$\boldsymbol\alpha$ (Ours)}} & \textbf{43.1} & \textbf{61.2} & \textbf{75.9} \\
\bottomrule
\end{tabular}
}
\label{tab:msfr}
\end{table}
\paragraph{Multi-scale Descriptor Head Design.}
Coarse-only and FPN-style serve as two straightforward baselines: the former relies solely on the coarsest resolution, while the latter adopts a standard top-down refinement without learning adaptive weights~\cite{lin2017feature}. 
Our Multi-Scale Fusion (MSF) framework introduces a structured multi-scale aggregation mechanism. 
Within MSF, the variant ``MSF --$\alpha$'' removes the predicted gating coefficient $\alpha$, while ``MSF +$\alpha$'' (Ours) incorporates $\alpha$ to adaptively weight contributions from different scales.

As shown in Table~\ref{tab:msfr}, Coarse-only and FPN-style both underperform, indicating that neither single-scale features nor uniform top-down fusion can provide robust representations under large viewpoint or geometric variations. 
MSF --$\alpha$ improves over FPN-style, confirming the benefit of structured multi-scale fusion; however, its lack of scale-adaptive weighting limits the ability to suppress noisy or unreliable scales. 
Introducing the $\alpha$-gated fusion in MSF +$\alpha$ consistently boosts performance across all angular thresholds, demonstrating that adaptive gating effectively selects informative scales while mitigating the influence of ambiguous ones. Equipping the fusion module with $\alpha$ gating leads to the most reliable and discriminative representations, yielding the best overall performance.

\paragraph{Ensemble Methods.}
We evaluate two levels of fusion: feature-level and match-level. 
For feature-level fusion, ``$\hat{P}$ as $F$'' replace the raw fine features $F$ in \cref{eq:warp} with the refined pyramid $\hat{P}$.
For match-level methods, the \emph{Warp head} and \emph{Descriptor head} operate independently, producing dense matching flow and descriptors respectively. 
On top of these two heads, we test two fusion strategies. 
\emph{Region guidance} assumes that the sparse correspondences from the descriptor head are highly reliable; it uses these descriptor-based matches as anchor points and samples the warp head’s dense flow field in local neighborhoods around these anchors to produce guided correspondences. 
The \emph{Confidence ensemble}, in contrast, performs fusion in a more principled manner. 
Instead of directly pooling all matches and selecting the globally highest-confidence ones, it operates \emph{within each head} by selecting only the high-confidence correspondences from the warp head and from the feature head separately. 
During merging, it further enforces a balanced number of matches from the two heads, preventing either the warp or descriptor branch from dominating the final set.

Table~\ref{tab:ensemble} shows that replacing $F$ with the refined $\hat{P}$ leads to weaker results on AerialMega, suggesting that refined high-resolution features are less stable under extreme viewpoint changes. 
Among match-level methods, using either head alone already provides competitive performance, with the warp head slightly stronger on outdoor and aerial benchmarks. 
Region guidance, however, fails to provide consistent improvements and degrades performance on aerial scenes, suggesting that assuming all descriptor matches are reliable leads to error propagation, especially in challenging wide-baseline scenarios. 
The Confidence ensemble achieves the best performance across all benchmarks, benefiting from head-wise confidence filtering and match-number balancing, which together produce a more stable and diverse correspondence set.

\begin{table}[t]
\centering
\caption{\textbf{Ablation on ensemble strategies.}
We evaluate different feature-level fusion and match-level ensembles on outdoor (MegaDepth), indoor (ScanNet), and aerial-to-ground (AerialMega) benchmarks at \emph{low resolution}, measured by AUC@5\degree. 
``$\hat{P}$ as $F$'' denotes replacing our fused pyramid $\hat{P}$ with the raw feature pyramid $F$.
Match-level methods include independent heads and their fusion variants.}
\resizebox{\linewidth}{!}{
\begin{tabular}{lcccc}
\toprule
\textbf{Method} & \textbf{MegaDepth} & \textbf{ScanNet} & \textbf{AerialMega} & \textbf{Mean} $\uparrow$ \\
\midrule
\multicolumn{5}{c}{\textit{Feature-level fusion}} \\
\midrule
$\hat{P}$ as $F$  
& 46.0 & 34.0 & 48.7 & 42.9  \\
\midrule
\multicolumn{5}{c}{\textit{Match-level methods}} \\
\midrule
Warp head only        
& 45.5 & 33.8 & 49.5 & 42.9 \\
Descriptor head only     
& 43.1 & 33.7 & 48.8 & 41.9  \\
Region guidance        
& 42.2 & 32.8 & 44.5 & 39.8 \\
\textbf{Confidence ensemble (Ours)} 
& \textbf{46.4} & \textbf{34.2} & \textbf{50.0} 
&  \textbf{43.5}\\
\bottomrule
\end{tabular}
}
\label{tab:ensemble}
\end{table}

\clearpage
{
    \small
    \bibliographystyle{ieeenat_fullname}
    \bibliography{main}
}


\end{document}